\documentclass[letterpaper]{article} 
\usepackage{aaai2026}  
\usepackage{times}  
\usepackage{helvet}  
\usepackage{courier}  
\usepackage[hyphens]{url}  
\usepackage{graphicx} 
\usepackage{natbib}  
\usepackage{caption} 
\usepackage{algorithm}
\usepackage{algorithmic}

\usepackage{multirow}
\usepackage{array} 
\usepackage{tabularx}
\usepackage{booktabs} 

\usepackage{amssymb}
\usepackage{amsmath}
\usepackage{algorithm}
\usepackage{algorithmic}
\usepackage{subfigure}
\usepackage{color}
\usepackage{colortbl}  
\usepackage[dvipsnames]{xcolor}
\usepackage{makecell}
\definecolor{c1}{RGB}{244,152,178}
\definecolor{c2}{RGB}{251,211,210}
\definecolor{c3}{RGB}{253,238,238}

\usepackage{newfloat}
\usepackage{listings}
\DeclareCaptionStyle{ruled}{labelfont=normalfont,labelsep=colon,strut=off} 
\floatstyle{ruled}
\newfloat{listing}{tb}{lst}{}
\floatname{listing}{Listing}
\title{Split-Layer: Enhancing Implicit Neural Representation by Maximizing the Dimensionality of Feature Space}
\author{
    Zhicheng Cai\textsuperscript{\rm 1},
    Hao Zhu\textsuperscript{\rm 1},
    Linsen Chen\textsuperscript{\rm 1},
    Qiu Shen\textsuperscript{\rm 1,\rm 2},
    Xun Cao\textsuperscript{\rm 1,\rm 2}\\
}
\affiliations{
    \textsuperscript{\rm 1}School of Electronic Science and Engineering, Nanjing University\\
    \textsuperscript{\rm 2}Key Laboratory of Optoelectronic Devices and Systems with Extreme Performances of MOE, Nanjing University\\
}

\usepackage{bibentry}

\begin{document}

\maketitle

\begin{abstract}
Implicit neural representation (INR) models signals as continuous functions using neural networks, offering efficient and differentiable optimization for inverse problems across diverse disciplines. However, the representational capacity of INR—defined by the range of functions the neural network can characterize—is inherently limited by the low-dimensional feature space in conventional multilayer perceptron (MLP) architectures. While widening the MLP can linearly increase feature space dimensionality, it also leads to a quadratic growth in computational and memory costs. To address this limitation, we propose the split-layer, a novel reformulation of MLP construction. The split-layer divides each layer into multiple parallel branches and integrates their outputs via Hadamard product, effectively constructing a high-degree polynomial space. This approach significantly enhances INR’s representational capacity by expanding the feature space dimensionality without incurring prohibitive computational overhead. Extensive experiments demonstrate that the split-layer substantially improves INR performance, surpassing existing methods across multiple tasks, including 2D image fitting, 2D CT reconstruction, 3D shape representation, and 5D novel view synthesis.
\end{abstract}

\section{Introduction}
Implicit Neural Representation (INR)~\cite{sitzmann2020implicit,liu2023finer} is reshaping the foundation of signal processing, garnering significant attention across various fields. Unlike traditional explicit discrete representations, INR leverages neural networks to implicitly model the continuous functional relationship between signal coordinates and their attributes, overcoming the limitations of traditional one in terms of storage efficiency and differentiability~\cite{martel2021acorn}. By integrating seamlessly with physical laws, INR provides a new paradigm for solving inverse problems in a self-supervised manner, thereby reducing the reliance on large-scale paired datasets~\cite{deng2009imagenet} in deep learning and mitigating the challenges of local minima~\cite{bubeck2015convex} in classical optimization problems. These unique advantages have positioned INR as a powerful tool with vast application potential, spanning domains such as computational mathematics~\cite{karniadakis2021physics,raissi2020hidden}, microscopic imaging~\cite{zhu2022dnf,zhou2023fourier,zhang2025single,zhou2024physics}, medical imaging~\cite{sun2024medical,shen2025cardiacfield,shen2024continuous}, and consumer-grade graphics rendering~\cite{mildenhall2021nerf,zhu2023pyramid}. As a result, INR not only offers new opportunities for research and development but also presents novel challenges that warrant further exploration.

However, the quality of inverse reconstruction is fundamentally determined by the representational capacity of INRs, for which the widely used multilayer perceptron (MLP) network suffers the well-known spectral-bias~\cite{rahaman2019spectral}. To enhance the capacity of INR, two categories of approaches are proposed. The first is mapping the low-dimensional input coordinates to high-dimensional manifolds using embeddings~\cite{tancik2020fourier,Chen2023TOG} or hash tables~\cite{muller2022instant,zhu2023disorder}.
Another is activating the neurons with specialized nonlinearity instead of the traditional ReLU~\cite{nair2010rectified}, such as periodic/variable-periodic~\cite{sitzmann2020implicit,liu2023finer} and wavelet~\cite{saragadam2023wire} activation functions.
However, these approaches primarily \textit{make certain features more easier to be learned by introducing a learning bias}~\cite{yuce2022structured} (e.g., Fourier bias and wavelet bias), rather than \textit{expanding the range of features that can be learned}.

\begin{figure*}[!ht]
  \centering
  \includegraphics[width=0.75\linewidth]{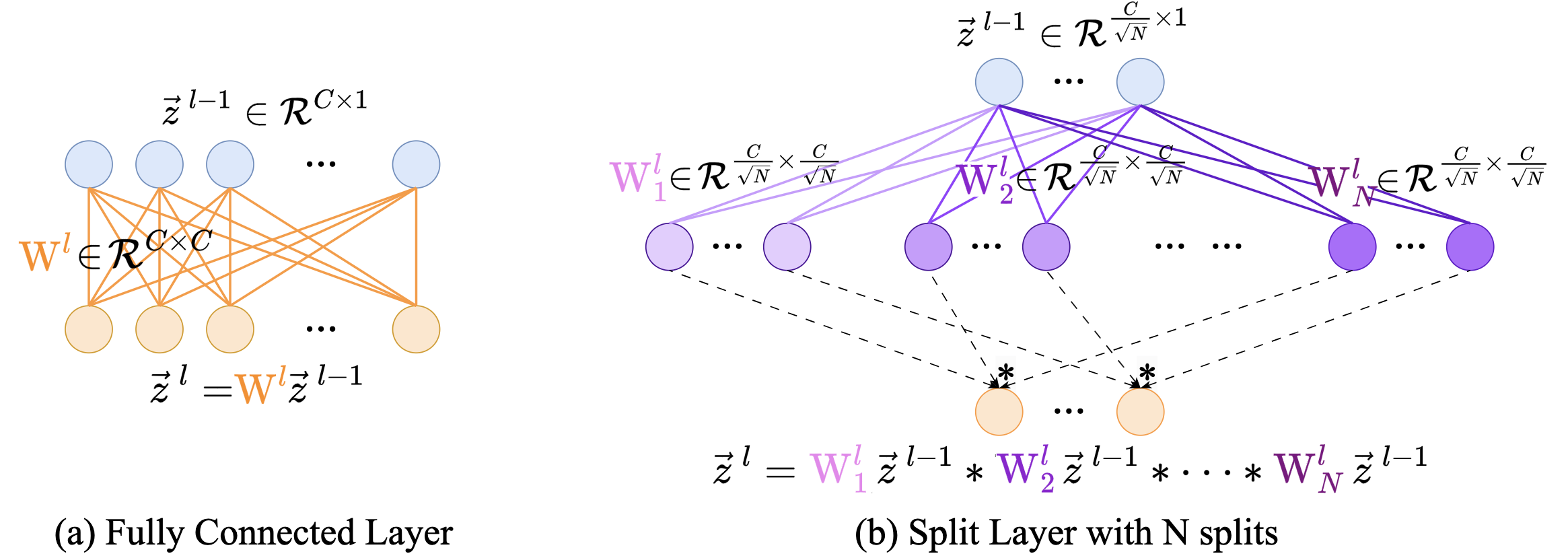}
  \caption{The diagram of split-layer. Solid lines represent learnable weights, and dashed lines represent the Hadamard product.}
  \label{fig:split}
\end{figure*}

Essentially, the representational capacity of INRs is defined by the diversity of features that they can represent, namely, the \textit{feature space}~\cite{scholkopf1999input,kansizoglou2021deep,zohdinasab2023efficient,ma2024rewrite}.
Specifically, for a fully connected layer with $C$ neurons, it spans an Euclidean space with $C$ dimensions as illustrated in Sec.~\ref{sec:fs}. 
Scaling up the layer width is a common and effective strategy to enhance representational capacity~\cite{hornik1989multilayer,kaplan2020scaling}, as it actually enlarges the feature space's dimensionality, which is proportional to the layer width.
However, linearly increasing the dimensionality results in a quadratic growth of parameters due to the formulation of MLPs, which incurs heavy computational budgets.

Such a problem is closely related to the improper fully connected mechanism designed in MLP, where the dimensionality of the feature space has a linear correction with number of neurons. By re-organizing the connection mechanism as a ``split''-style (see Fig.~\ref{fig:split}), the dimensionality of features space could be significantly increased (Fig.~\ref{fig:visf}). Thus, we propose the split-layer, where the original fully connected layer is split into $N$ parts, each with an individual weight matrix $\mathbf{W}^n\in\mathbb{R}^{\frac{C}{\sqrt{N}}\times \frac{C}{\sqrt{N}}}$, and then \textit{integrates} the outputs of these parts though the Hadamard product, thus formulating high-degree polynomials. As theoretically demonstrated in Sec.\ref{sec:fs}, this modification can effectively expand the dimensionality of the feature space from $C$ to an extremely large dimensionality of $\binom{\frac{C}{\sqrt{N}} + N - 1}{N}$ with no additional parameters or computational costs  (Fig.\ref{fig:visf} visualizes the feature space for a simple case), significantly enhancing the INR representational capacity (improving the performance up to $45\%$ in image fitting task).

Specifically, we make the following contributions,
\begin{enumerate}
    \item We propose the \textit{split-layer}, which is a universal tool for enhancing the representational capacity of INR with different backbones.
    \item We theoretically demonstrate the improvements of split-layer on extending the feature space's dimensionality and give the law for deriving the optimal split.
    \item We substantiate the significant improvements of split-layer for different INR backbones (MLP, SIREN, Gauss, PEMLP, WIRE, and FINER) and various tasks (2D image representation, 2D computed tomography reconstruction, 3D shape representation, and 5D novel view synthesis.).
\end{enumerate}

\section{Related Work}
\subsection{Implicit Neural Representations}
Implicit Neural Representations (INRs), also referred to as coordinate networks or neural fields~\cite{tancik2020fourier,sitzmann2020implicit,cai2024batch}, are parameterized by multi-layer perceptron (MLP) models to learn mappings from input coordinates to signal attributes, and are gradually emerging as the dominant paradigm for multi-modal signal representation and inverse problem optimization.

Due to the powerful approximation capabilities of neural networks, INRs offer advantages in compact modeling and efficient storage, enabling their successful applications in various signal representation and compression tasks, such as 1D audio~\cite{gao2022objectfolder}, 2D images~\cite{strumpler2022implicit}, 3D videos/structures/spectrum~\cite{shi2025quantizing,lindell2022bacon,shi2024hiner}, 4D light fields~\cite{sitzmann2021light}, and 5D radiance fields~\cite{mildenhall2021nerf}.
In addition, due to the continuity and differentiability of neural networks, INRs can seamlessly integrate various physical processes to solve partial differential equations, providing an promising paradigm for addressing a wide range of inverse problems, especially with limited measurements, including inverse rendering for novel view synthesis~\cite{mildenhall2021nerf,zhu2023pyramid}, phase recovery in lensless imaging~
\cite{zhu2022dnf}, 3D static and 4D dynamic heart reconstruction from 2D ultrasound measurements~\cite{shen2025cardiacfield,shen2024continuous} and so on~\cite{park2019deepsdf,chabra2020deep,molaei2023implicit,gupta2022neural,reed2021dynamic,xie2022neural}. 

\subsection{Improving the Representational Capacity of INR}
The representational capacity of INRs plays a critical role in determining the quality of reconstruction and rendering. Existing methods for enhancing the INR representational capacity can be broadly classified into two categories. 
The first is mapping the low-dimensional input coordinates to high-dimensional manifolds using embeddings or hash tables. PEMLP\cite{tancik2020fourier} encodes the coordinates with Fourier basis embeddings. Poly-INR~\cite{Singh_2023_CVPR} and PINs~\cite{landgraf2022pins} propose polynomial basis embeddings. DINER~\cite{zhu2023disorder,zhu2023rhino} maps the coordinates with single scale full-resolution hash table. InstantNGP~\cite{muller2022instant} maps with multi-scale pyramid hash tables. PIXEL~\cite{kang2022pixel} uses multiple shifting hash tables. 
Another approach involves activating the hidden neurons with specialized nonlinearity instead of the traditional ReLU. SIREN~\cite{sitzmann2020implicit} activates with periodical sinusoidal function, while GARF~\cite{chng2022gaussian} introduces the non-periodic Guassian activation. FINER~\cite{liu2023finer} proposes the variable-periodic activation.
WIRE~\cite{saragadam2023wire} employs wavelet transformations as activation.
MFN~\cite{fathony2020multiplicative} and Bacon~\cite{lindell2022bacon} activate with compositing multiple filters.

However, current INRs still predominantly rely on the vanilla MLP formulated with fully connected layers, which inherently limits their representational capacity, namely, the dimensionality of feature space is linear to the width of the network. To address this limitation, we propose split-layer, which reformulates the model with high-degree forms and significantly enlarge the feature space, thus enhancing the representational capacity of INRs from model formulation perspective. 

\section{Split-Layer for INR}
In this section, we first elucidate the formulation of INR and the relationship between the feature space and the representational capacity of INR. Then, the split-layer is proposed and analysed on how it modifies the feature space and INR's representational capacity.

\subsection{Formulation of INR}
Given a signal $\{(\Vec{x}_i,\Vec{y}_i)\}_{i=1}^{K}$, INR takes its coordinate $\Vec{x}_{i}=[{x}_{i}^{1},{x}_{i}^{2},...,{x}_{i}^{d_{in}}]^{\top}$ as input and outputs the corresponding attributes $\Vec{y}_{i}=[{y}_{i}^{1},{y}_{i}^{2},...,{y}_{i}^{d_{out}}]^{\top}$, where $d_{in}$ and $d_{out}$ are the dimensions of input coordinate and output attribute, $K$ is the length of the signal. Mostly, INR is built upon the multilayer perceptron (MLP), that
\begin{equation}
\begin{aligned}
\Vec{z}^{\:0} &= \Vec{x}\\
\Vec{z}^{\:l}&=\mathbf{W}^l\Vec{z}^{\:l-1}+\Vec{b}^{\:l-1},\ \  l=1\dots L \\
\Vec{z}^{\:l}_{act}&=\rho(\Vec{z}^{\:l})\\
f(\Vec{x};\theta) &=  \Vec{z}^{\:L}_{act}
\label{siren}
\end{aligned}
\end{equation}
where $L$ is the number of layers, $\Vec{z}^{\:l}$ is the linear combination of the $l-1$-th layer, and $\Vec{z}^{\:l}_{act}$ is the output of $l$-th layer, $\rho$ is the activation function\cite{sitzmann2020implicit,liu2023finer}, $\mathbf{W}^{l}$ is the weight matrix and $\Vec{b}^{\:l-1}$ is the bias vector, $\theta$ refers to the network parameters to be optimized. For the sake of derivation, the width of each layer is defined as $C$, as a result, $\mathbf{W}^{l}\in \mathbb{R}^{C\times C}$ and $\Vec{z}^{\:l}=\{z_1^l,z_2^l,...,z_C^l\}$.

\subsection{Feature space and representational capacity of INR}~\label{sec:fs}
According to the definition in Eqn.\ref{siren}, the $i$-th element $z_i^{\:l}$ output by $l$-th layer is 
\begin{equation}
z_i^l = w_{i1}z_1^{l-1} +  w_{i2}z_2^{l-1} + \cdot\cdot\cdot + w_{iC}z_C^{l-1},
\label{fs-mlp}
\end{equation}
where $w_{ij}$ refers to the $i,j$-th element of $\mathbf{W}$. Note that the bias term $\Vec{b}^{\:l-1}$ is removed here to facilitate derivation. From this equation, the output is a composition of $C$ individual and linearly independent elements $\{z_1^{l-1},z_2^{l-1},\cdot\cdot\cdot,z_C^{l-1}\}$, thus constituting the basis that spans the \textit{feature space}\cite{bengio2013representation} $\mathcal{F}^{C}=\mathcal{S}(z_1^{l-1},z_2^{l-1},\cdot\cdot\cdot,z_C^{l-1})$, which is a $C\!-\!dimensional$ \textit{Euclidean space}.

According to recent study~\cite{yuce2022structured,benbarka2022seeing,zhu2024finer++}, the representational capacity of INR could be viewed as a linear combination of Fourier (or Gauss/Wavelet) bases, 
\begin{equation}
\begin{aligned}
f(\mathbf{x};\theta) \in \left\{ \sum_{\omega'\in\mathcal{H}^C_{\omega}} c_{\omega'}\sin(\omega'\mathbf{x}+\phi_{\omega'})\ |\ c_{\omega'},\phi_{\omega'} \in Q \right\}
\label{siren-freq}
\end{aligned}
\end{equation}
where $Q$ is the set of rational number, $\mathcal{H}^C_{\omega}$ is the supported frequency set whose size is proportional to the number of linearly independent elements in feature space.

As a result, widening the network layer, for example with a coefficient of $2$, will expand the feature space from $\mathcal{F}^{C}$ to $\mathcal{F}^{2C}$ and thus enhancing the representational capacity of INR, but at the cost of significantly improved computational budget (from $C^2$ to $4C^2$).

\subsection{Split-layer reformulates the feature space of INR}
To expand the feature space to high dimensionality without increasing the number of parameters and computational budget, we propose the \textit{split-layer} to reformulate the MLP with high-degree forms. As shown in Fig.~\ref{fig:split}, split-layer splits each layer as several individual parts and then integrates the outputs of these branches through the Hadamard product. Supposing the $l$-th layer is split into $N$ parts (as shown in Fig.~\ref{fig:split}b, the neurons of each part is set as $\frac{C}{\sqrt{N}}$ to maintain the total parameters is equivalent to $C^2$, thus $N$ ranges from 2 to $C^2$. Correspondingly, the weight matrices of these $N$ branches are denoted as $\mathbf{W}_1^{l},\mathbf{W}_2^{l},\cdot\cdot\cdot,\mathbf{W}_N^{l}\in\mathbb{R}^{\frac{C}{\sqrt{N}}\times \frac{C}{\sqrt{N}}}$, and the input becomes $\Vec{z}^{\:l-1}=[z_1^{\:l-1},z_2^{\:l-1},\cdot\cdot\cdot,z_\frac{C}{\sqrt{N}}^{\:l-1}]$. In this case, the split-layer with $N$ splits can be formulated as $\Vec{z}^{\:l}=\mathbf{W}_1^{l}\Vec{z}^{\:l-1}\ast\mathbf{W}_2^{l}\Vec{z}^{\:l-1}\ast\cdot\cdot\cdot\ast\mathbf{W}_N^{l}\Vec{z}^{\:l-1}$, where $\ast$ is the Hadamard product operation, referring to the element-wise product between vectors. In detail, the $i$-th element $z_{i}^{l}$ of $\Vec{z}^{\:l}$ could be written as,
\begin{equation}
\begin{aligned}
z_i^{l} &= \sum^{\frac{C}{\sqrt{N}}}_{j=1} w^1_{ij}z_{j}^{l-1}\ast\sum^{\frac{C}{\sqrt{N}}}_{j=1} w^2_{ij}z_{j}^{l-1}\ast\cdot\cdot\cdot\ast\sum^{\frac{C}{\sqrt{N}}}_{j=1}w^N_{ij}z_{j}^{l-1} \\
&=\prod_{n=1}^{N} \left( \sum_{j=1}^{\frac{c}{\sqrt{N}}} w^n_{ij} z_j^{l-1} \right)=\sum_{j_1=1}^{\frac{c}{\sqrt{N}}} \cdots \sum_{j_N=1}^{\frac{c}{\sqrt{N}}} 
\prod_{n=1}^{N} w_{ij_n}^n \prod_{n=1}^{N} z_{j_n}^{l-1}\\
&=\sum_{(j_1, j_2, \ldots, j_N)} 
\left( \prod_{n=1}^{N} w_{ij_n}^n \right) 
\left(z_{j_1}^{l-1}z_{j_2}^{l-1}\cdots z_{j_N}^{l-1} \right)
\label{splitN-mlp}
\end{aligned}
\end{equation}
where $w_{ij}^{n}$ refers to the $i,j$-th element of the $n$-th weight matrix $\mathbf{W}_n^{l}$. Our goal is to calculate the number of different items obtained by the \textit{N-degree symmetric homogeneous polynomials} $z_{j_1}^{l-1}z_{j_2}^{l-1}\cdots z_{j_N}^{l-1}$, which is equivalent to selecting all combinations of $N$ elements from the set $\{x_1,x_2,\cdot\cdot\cdot,x_\frac{C}{\sqrt{N}}\}$ in a reproducible and order-independent manner. As a consequence, the total number of distinct items composing $z_i^{l}$ is described by the \textit{binomial coefficient} $\binom{\frac{C}{\sqrt{N}} + N - 1}{N}$.
These homogeneous polynomials are linearly independent to each other, forming the basis that spans a $\binom{\frac{C}{\sqrt{N}} + N - 1}{N}\!-\!dimensional$ \textit{feature space} $\mathcal{F}^{\binom{\frac{C}{\sqrt{N}} + N - 1}{N}}$.
In this way, we obtain an extremely large feature space of $\binom{\frac{C}{\sqrt{N}} + N - 1}{N}$ dimensions with the same computations.

\begin{figure}[!t]
  \centering
  \includegraphics[width=\linewidth]{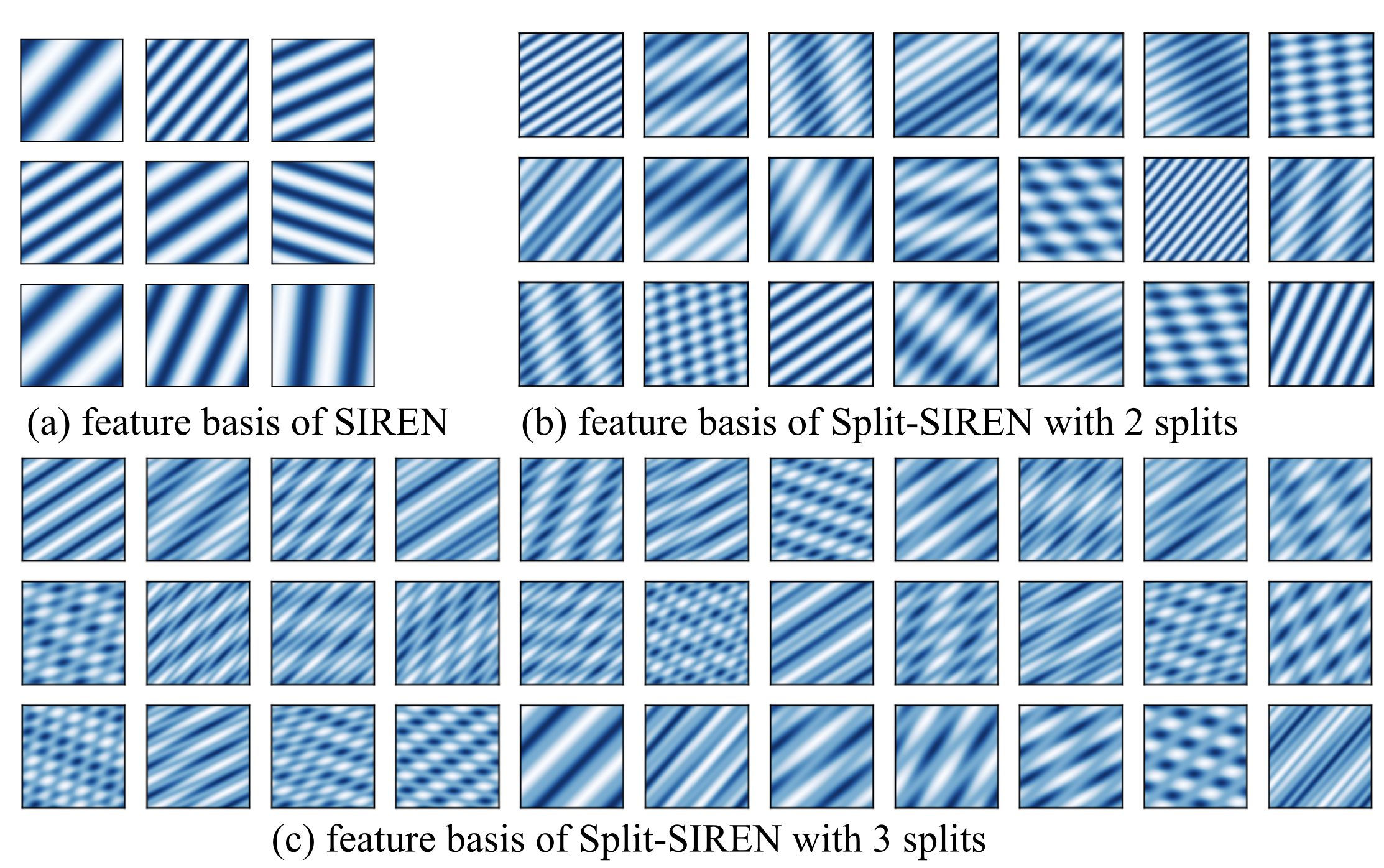}
  \caption{We visualize the features of SIREN with 9 hidden neurons (a), and the corresponding Split-SIREN with 2 splits (b), and corresponding Split-SIREN with 3 splits (c) on a 2D image fitting task. As can be observed, split-layer introduces more diverse feature basis, significantly enlarging the feature space of the original model.}
  \label{fig:visf}
\end{figure}

\subsection{Discussion}
\noindent\textbf{Feature Comparisons between MLP with Different Splits}. As analyzed mentioned above, split-layer reformulates the construction of MLP and extends the feature space without adding network parameters. To verify this issue, we apply the split-layer to the widely used INR architecture, \textit{i.e.}, the SIREN network~\cite{sitzmann2020implicit}.  Fig.~\ref{fig:visf} compares the features output by the first layer of the SIREN. In Fig.~\ref{fig:visf}(a), the original SIREN only provides 9 features where each feature contains only one frequency. By applying the split-layer to the SIREN, it is noticed that more features with complex distributions are produced (as shown in Fig.~\ref{fig:visf}(b) and (c)), verifying the effectiveness of the proposed split-layer in extending the feature space.

\begin{figure}[!t]
  \centering
  \begin{subfigure}
    \centering
    \includegraphics[width=0.48\linewidth]{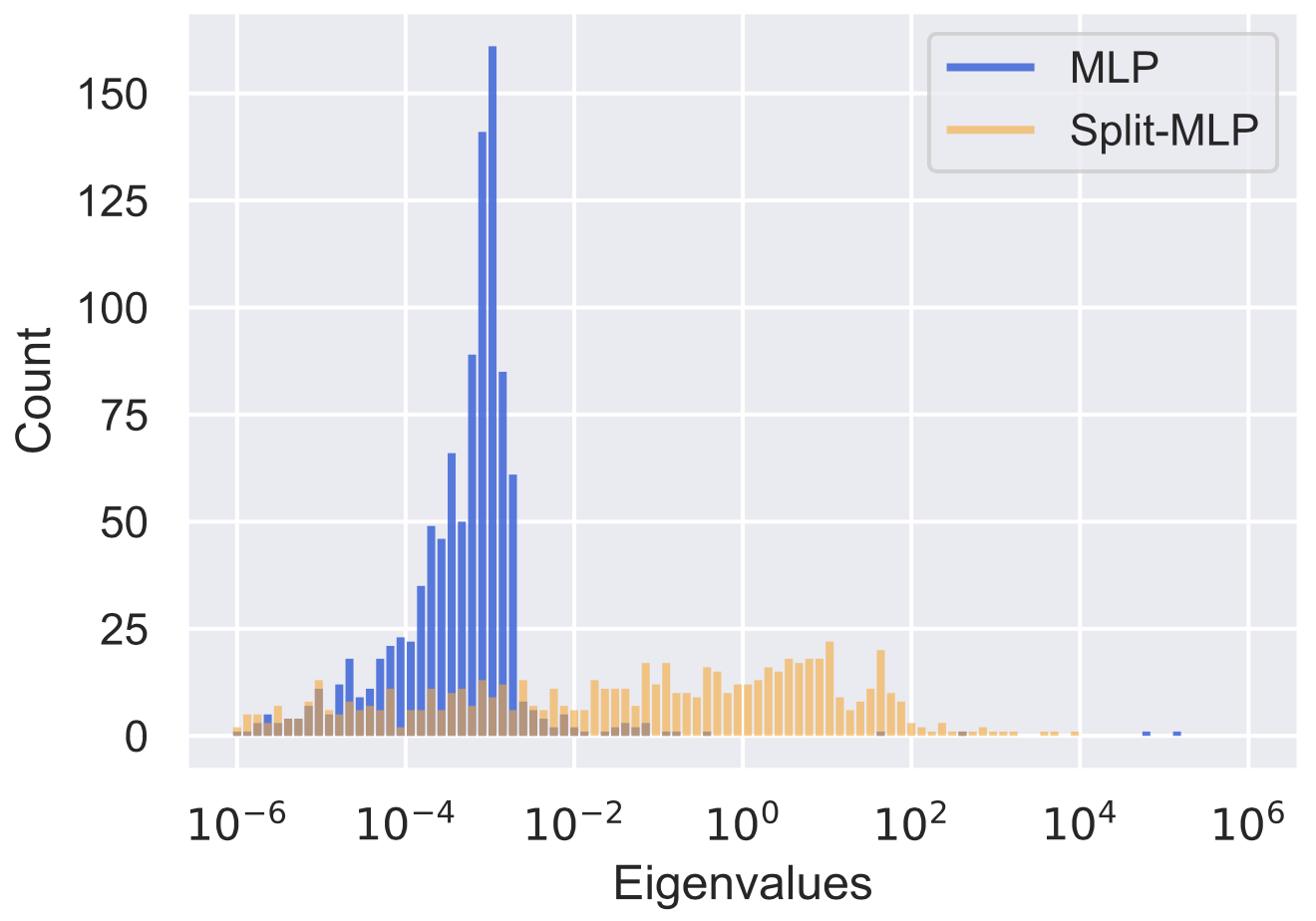}
  \end{subfigure}
  \hfill
  \begin{subfigure}
    \centering
    \includegraphics[width=0.48\linewidth]{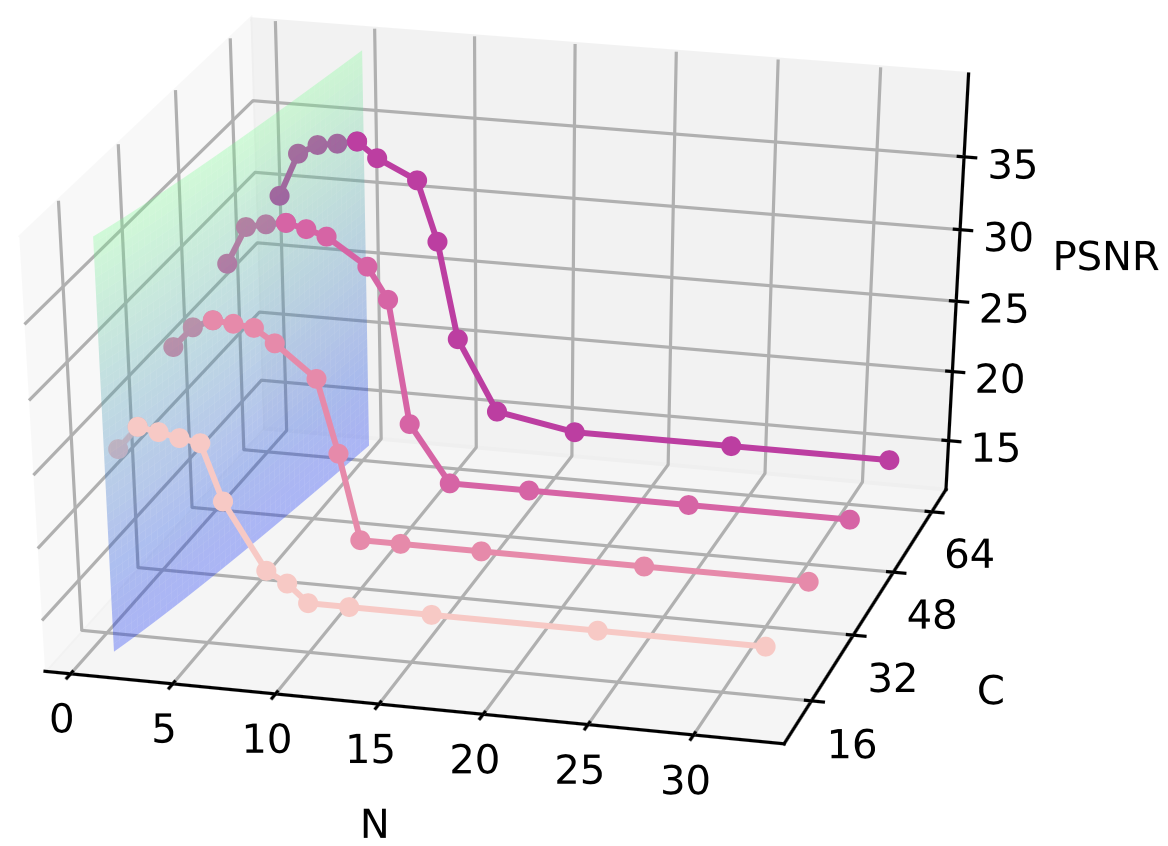}
  \end{subfigure}
  \vspace{-1em} 
  \caption{Left: Comparisons of the distribution of NTK's eigenvalues. Most of the MLP's eigenvalues are smaller than 1, while the eigenvalues of Split-MLP are increased to the range of $[10^{-2},10^2]$, meaning better performance of Split-MLP for representing a signal with high-frequency components. Right: Verification of the Eqn.~\ref{eopt-n} for obtaining the optimal split. The curved surface visualizes the Eqn.~\ref{eopt-n} and four curves plots the quality of image fitting with different splits. It is observed that the optimal result of each curve appears near the surface, verifying the robustness of Eqn.~\ref{eopt-n}.}
  \vspace{-1em} 
  \label{fig3}
\end{figure}

\noindent\textbf{Neural Tangent Kernel Perspective}.
Neural tangent kernel (NTK)~\cite{jacot2018neural,tancik2020fourier}, which approximates the training of neural network as kernel regression, has become a popular lens for monitoring the dynamic behaviors and convergence of a neural network. Previous studies~\cite{tancik2020fourier,cai2024batch,cai2024towards,liu2023finer} theoretically demonstrate that more larger eigenvalues of NTK leads to faster convergence for learning high-frequency components.
Fig.~\ref{fig3} left visualizes the distributions of the NTK's eigenvalues of MLP and Split-MLP (with 2 splits) for learning a 1D signal.
As can be observed, the NTK's eigenvalues of MLP exhibit the pathological distribution~\cite{cai2024batch,cai2024towards}, mostly concentrated around $10^{-3}$ and close to $0$, with almost no large eigenvalues, thereby limiting the convergence for the high-frequency components. In contrast, the Split-MLP shows a more even distribution of NTK's eigenvalues, significantly expanding the distribution of NTK eigenvalues to the range of $[10^{-2}, 10^2]$, offering better convergence performance than the original MLP, thus validating the effectiveness of split-layer form the neural tangent kernel perspective.

\noindent\textbf{Optimal Split}.
According to the derivation of the feature space, improving the split $N$ will increase the dimensionality of the feature space at the cost of reducing the size of each weight matrix $\{\mathbf{W}_{i}^l\}_{i=1}^{N}$. On the contrary, a smaller weight matrix has less freedom to explore different combinations of characteristics and hierarchical structures~\cite{unterthiner2020predicting,navon2023equivariant,denil2013predicting}. We empirically find that the optimal split could be written as,
\begin{equation}
    N^*\approx (0.17C)^{\frac{2}{3}}
\label{eopt-n}
\end{equation}
To verify the robustness of the selection of $N^*$, we conduct the 2D image fitting task with different settings of network width $C$. In Fig.~\ref{fig3} right, the curved surface indicates Eqn.~\ref{eopt-n} and the four curves indicate results with different network widths. It is observed that the optimal node of each curve always appears near the surface, verifying the robustness of the Eqn.~\ref{eopt-n}.


\begin{table*}[!t]
\centering
\begin{tabular}{l|l|cccccc} 
\toprule
Task & Method                                 &  ReLU  & SIREN  & Gauss  & PEMLP   & WIRE  & FINER\\
\midrule
\multirow{3}{*}{\makecell{Image Fitting\\PSNR (dB$\uparrow$)}}  & Baseline    & 21.24  & 38.52  & 31.74  & 29.60   & 31.31 & 38.72\\
                                & Split       & 30.89  & 39.25  & 40.84  & 40.78   & 35.39 & 40.09\\
    & Gain   &$\uparrow$45.43\%&$\uparrow$1.90\% &$\uparrow$28.67\%&$\uparrow$37.77\% &$\uparrow$13.03\%&$\uparrow$3.54\% \\
\midrule
\multirow{3}{*}{\makecell{CT Reconstruction\\PSNR (dB$\uparrow$)}}&Baseline   & 26.78  & 18.32  & 27.44  & 28.11   & 28.26 & 25.10\\
                                 &Split       & 30.19  & 29.11  & 28.72  & 32.29   & 31.67 & 29.58\\
    & Gain  &$\uparrow$12.73\%&$\uparrow$58.90\%&$\uparrow$4.66\% &$\uparrow$14.87\% &$\uparrow$12.07\%&$\uparrow$17.84\%\\
                                 \midrule
\multirow{3}{*}{\makecell{Shape Representation\\Chamfer distance ($\downarrow$)}}&Baseline& 1.00e-4 & 2.44e-5 & 2.19e-5 & 9.76e-6 & 1.38e-5 & 4.89e-5\\
                                 &Split       & 2.01e-5 & 1.21e-5 & 5.33e-6 & 4.59e-6 & 9.41e-6 & 1.82e-5\\
    &  Gain
&$\uparrow$79.90\%&$\uparrow$50.41\%&$\uparrow$75.66\%&$\uparrow$52.97\%&$\uparrow$31.81\% & $\uparrow$62.78\%\\
\bottomrule
\end{tabular}
\caption{Results of different INRs on image fitting task (measured in PSNR), CT reconstruction task (measured in PSNR), and shape representation task (measured in chamfer distance error, lower chamfer distance means better performance).}
\label{exp}
\end{table*}

\begin{figure*}[!t]
  \centering
  \includegraphics[width=0.95\linewidth]{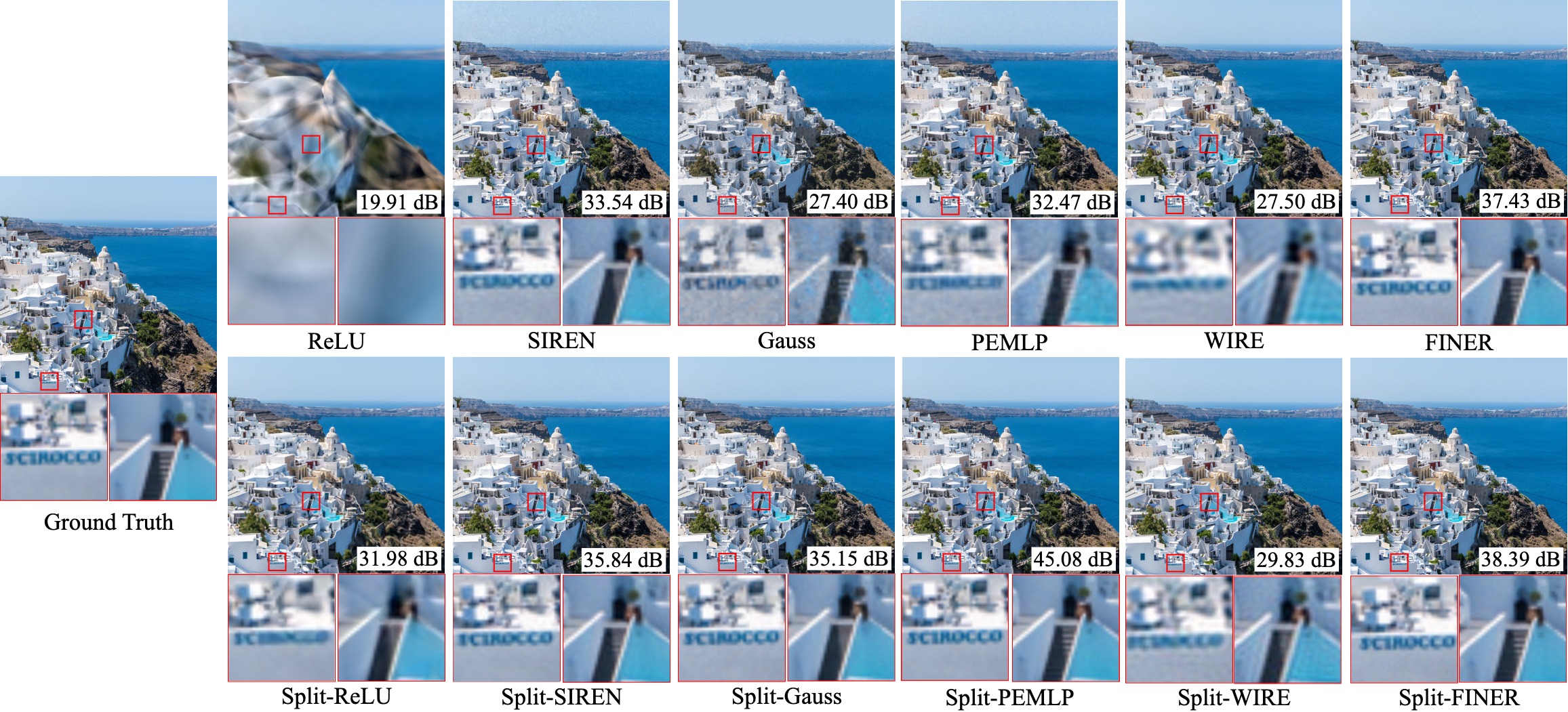}
  \caption{Comparisons of different methods for representing the 2D Image Santorini.}
  \label{fig:img}
  \vspace{-1em}
\end{figure*}

\section{Experiments}
\textbf{Tasks.} 
We validate the effectiveness of applying split-layer to INRs on four separate tasks, \textit{i.e.}, 2D image fitting, 2D computed tomography (CT) reconstruction, 3D shape representation, and 5D novel view synthesis. Note that the first 3 tasks are the ones widely used in INR research to evaluate performance, assessing the fundamental models in current INR studies, while the last task represents the main practical uses of INR and needs to be compared with a range of state-of-the-art methods.

\textbf{Compared methods.} 
For these three tasks, a total of twelve methods are compared, including the MLP with traditional ReLU (marked as ReLU)~\cite{nair2010rectified}, MLP with Fourier positional encoding (PEMLP)~\cite{tancik2020fourier},  SIREN~\cite{sitzmann2020implicit}, MLP with Gaussian activation (Gauss)~\cite{ramasinghe2022beyond}, WIRE~\cite{saragadam2023wire} and FINER~\cite{liu2023finer} .
As a common practice, the encoding scale of PEMLP is set as $10$~\cite{mildenhall2021nerf}, the frequency parameter $\omega$  of SIREN is set as $30$~\cite{sitzmann2020implicit}.
We replace all the original hidden fully connected layers of these six methods with split-layers to test the effectiveness of split-layer.
To facilitate deployment and more effectively illustrate the efficacy of our method, we consistently set the split number as $2$, which still yields excellent results.
The weights of all the networks are randomly initialized. For SIREN and FINER, we use the specific weight initialization schemes as raised in \cite{sitzmann2020implicit,liu2023finer}. For the left methods, we use the default LeCun initialization~\cite{lecun2002efficient}.

\begin{figure*}[t]
  \centering
  \includegraphics[width=0.95\linewidth]{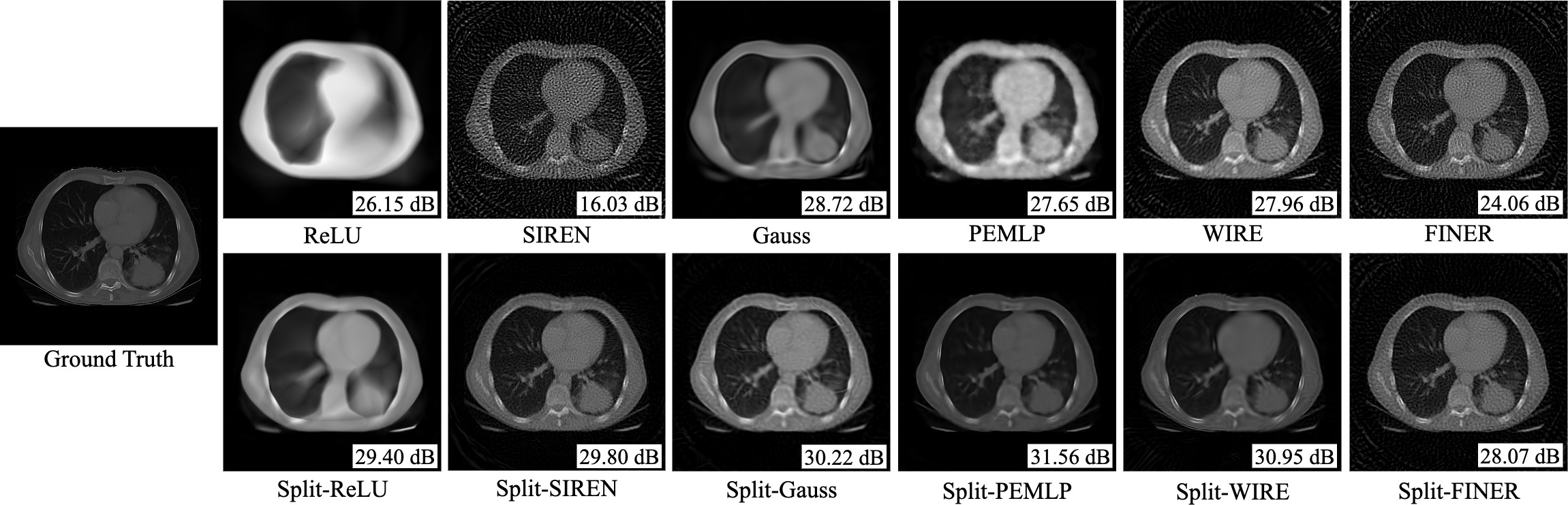}
  \caption{Comparisons of different methods for CT reconstruction. The corresponding error maps are also visualized.}
  \label{ct-vis}
  \vspace{-1em}
\end{figure*}

\subsection{2D Image Fitting}
\textbf{Configurations.}
We first use an image representation task to evaluate the performance of applying split-layer to different implicit neural representations. 
The INR aims at learning a mapping from 2D pixel location to 3D RGB color. 
We perform experiments on the \textit{Natural} image dataset~\cite{tancik2020fourier} consisting of 16 $512\times512$ RGB images.
We use networks with 4 hidden layers and a width $C$ of 256, which is sufficiently wide. 
We use the $L_2$ distance between the network output and the ground truth as the loss function. 
All the models are trained for 50,000 iterations using Adam optimizer~\cite{kingma2014adam}.

\textbf{Results.}
Tab.~\ref{exp} exhibits the average PSNR (Peak Signal-to-Noise Ratio) of these methods. 
As can be observed, split-layer significantly enhances the performance of ReLU by $9.65$dB, achieving a substantial improvement up to $45.4\%$.
Split-layer also significantly enhances the performance of PEMLP and Gauss by $11.18$dB and $9.10$, up to a remarkable improvement of $37.77\%$ and $28.67\%$, respectively. 
The experimental results validate that split-layer can effectively improve the representational capacity of different implicit neural representation models.

\textbf{Visualization.}
We visualize the reconstructed images by these methods in Fig.~\ref{fig:img}.
As can be observed, the reconstruction of ReLU is over-smoothed and poor-quality.
For PEMLP, the reconstructed image is blurry and fails to restore finer details.
For Gauss and SIREN, the reconstructed images exhibit obvious noise.
In contrast, applying split-layer significantly improves the reconstruction quality, with less noise and finer details clearly preserved (as shown by the characters and stages in the zoomed-in read box). Notably, Split-PEMLP achieves a high PSNR of $45.08$dB, demonstrating excellent reconstruction.
These results underscore the superiority of split-layer over existing methods

\subsection{2D Computed Tomography}
\textbf{Configurations.}
In the CT reconstruction task, we observe integral projections of a density field instead of direct supervisions. We train a network that takes in 2D pixel coordinates and predicts the corresponding volume density.
We conduct the experiments on the x-ray colorectal dataset~\cite{saragadam2023wire,clark2013cancer}, each image has a resolution of $512\times 512$ and is emulated with 100 CT measurements.
We use networks with 2 hidden layers and a width $C$ of 256. 
To solve the inverse problem, the network is indirectly supervised by the MSE loss between a sparse set of ground-truth integral projections and integral projections computed from the network's output.
All the models are trained for 5,000 iterations using Adam optimizer.

\textbf{Results.}
Tab.~\ref{exp} exhibits the average PSNR of these methods. 
As can be observed, split-layer significantly enhances the performance of ReLU by $3.41$dB, delivering an improvement up to $12.73\%$.
Split-layer also significantly boosts the performance of PEMLP and Gauss by $4.18$dB and $1.28$dB, achieving an improvement up to $14.87\%$ and $4.66\%$, respectively. 
Notably, the performance of original SIREN is very poor, which achieves a PSNR of only $18.32$dB. When applied with split-layer, the performance of SIREN is significantly enhanced up to $29.11$dB, bringing a remarkable improvement of $58.90\%$.
The experimental results effectively validate that split-layer can generally improve the representational capacity of implicit neural representations.

\textbf{Visualization}
We visualize the reconstructed CT images by these eight methods in Fig.~\ref{ct-vis},  the corresponding error maps are also visualized in the bottom right corner of each method.
As can be observed, ReLU leads to excessively smooth results, displaying only blurry patterns. The result of PEMLP is noisy, failing to exhibit the precise details. However, when split-layer is applied to ReLU and PEMLP, the reconstruction quality improves significantly, with finer details effectively represented.
Additionally, SIREN yields an extremely low PSNR of only $16$dB, likely due to the pre-encoded frequencies that introduce a bias inconsistent with the target, resulting in substantial noise. After applying the split-layer to SIREN, the reconstruction quality improves markedly, with the PSNR increasing to $29.80$dB.

\begin{table*}[!t]
\centering
\begin{tabular}{l|cccccccc|c} 
\toprule
Method     & Chair & Drums & Ficus & Hotdog& Lego  & Materials & Mic   & Ship  & Average \\ 
\toprule
NeRF       & 31.37 & 24.50 & 28.90 & 34.94 & 30.71 & 28.60     & 28.99 & 27.27 & 29.41  \\ 
Split-NeRF & 31.78 & 24.81 & 29.34 & 35.33 & 31.76 & 28.87     & 31.85 & 27.83 & 30.20  \\ 
\toprule
DVGO       & 34.07 & 25.30 & 32.59 & \cellcolor{c3}{36.75} & 34.65 & \cellcolor{c2}{29.59}     & 33.12 & 29.00 & 31.88   \\
Split-DVGO & \cellcolor{c3}{34.28} & \cellcolor{c1}{25.49} & \cellcolor{c3}{32.92} & \cellcolor{c2}{36.77} & \cellcolor{c2}{34.97} & \cellcolor{c1}{29.76}     & \cellcolor{c3}{33.35} & \cellcolor{c2}{29.27} & \cellcolor{c2}{32.10} \\ 
\toprule
DINER      & \cellcolor{c2}{34.49} & \cellcolor{c3}{25.43} & \cellcolor{c2}{33.28} & 36.45 & \cellcolor{c3}{34.82} & \cellcolor{c3}{29.58}     & \cellcolor{c2}{33.43} & \cellcolor{c3}{29.25} & \cellcolor{c3}{32.09}   \\ 
Split-DINER& \cellcolor{c1}{34.85} & \cellcolor{c2}{25.47} & \cellcolor{c1}{33.39} & \cellcolor{c1}{36.92} & \cellcolor{c1}{35.14} & \cellcolor{c2}{29.59}     & \cellcolor{c1}{34.01} & \cellcolor{c1}{29.49} & \cellcolor{c1}{32.36} \\ 
\bottomrule
\end{tabular}
\caption{Results of different methods on inverse rendering for novel view synthesis, measured in PSNR. We color code each cell as \colorbox{c1}{best}, \colorbox{c2}{second best}, and \colorbox{c3}{third best}.}%
\label{exp-nerf}
\end{table*}

\begin{figure*}[!ht]
  \centering
  \includegraphics[width=0.95\linewidth]{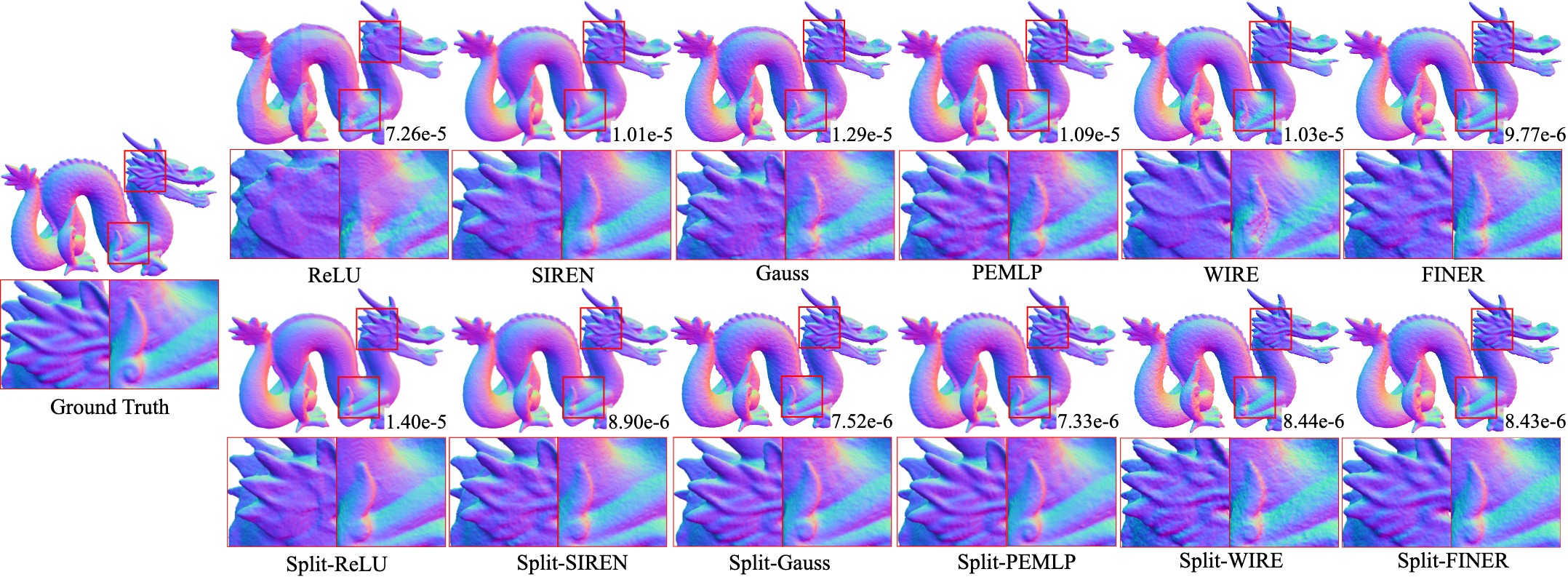}
  \caption{Meshes of Dragon generated with occupancy volumes by various methods.}
  \label{sdf-vis}
  \vspace{-1em}
\end{figure*}

\subsection{3D Shape Representation}
\textbf{Configurations.}
In this section, we demonstrate the representational capacity of INRs for representing 3D shapes as occupancy networks. 
To be specific, the input data is a mesh grid with $512^3$ resolution, where the voxels inside the volume are assigned as $1$, and the voxels outside the volume are assigned as $0$.
Then we use the occupancy network to implicitly represent a 3D shape as the “decision boundary” of INR, which is trained to output 0 for points outside the shape and 1 for points inside the shape. Test error is calculated using cross-entropy loss between the network output and the ground truth points.
We conduct the experiments on Stanford 3D Scanning Repository~\cite{standord-3D-scanning}.
We use networks with 2 hidden layers and a width $C$ of 256. 
All the models are trained for 200 epochs using Adam optimizer. 
200,000 points are randomly sampled in each iteration during the training process. The network outputs are extracted as a $512^3$ grid using marching cubes~\cite{lorensen1998marching} with a threshold of 0.5 for evaluation and visualization.

\textbf{Results.}
Tab.~\ref{exp} exhibits the experimental results evaluated chamfer distance (lower chamfer distance means better performance).
As observed, split-layer markedly improves the performance of ReLU, SIREN, Gauss, and PEMLP by $79.90\%$, $50.41\%$, $75.66\%$, $52.97\%$  on average, demonstrating the general effectiveness of split-layer. Moreover, Split-PEMLP obtains the best performance, achieving a exceedingly low CD of $4.59e\!-\!6$ (close to the optimal value $0$).
The experimental results validate the general effectiveness of split-layer.

\textbf{Visualization.}
Fig.~\ref{sdf-vis} visualizes the meshes of thai-statue scene represented by these eight methods.
As observed, ReLU makes the surface over-smooth, while split-layer significantly improves such phenomenon allowing finer details to emerge. 
Besides, SIREN and Gauss introduces too many undesired textures and fluctuations on the surface, indicating over-fitting to high-frequencies and noise.
Furthermore, Split-PEMLP achieves the best representational result, providing clear details without incurring artifacts or noise.  

\subsection{Inverse Rendering for Novel View Synthesis}
\textbf{Configurations.}
We further demonstrate the effectiveness of split-layer on inverse rendering for novel view synthesis using the neural radiance fields (NeRF)~\cite{mildenhall2021nerf}. NeRF models the 3D world as a 5D radiance fields using INRs, where the input contains the 3D position and 2D viewing direction of a point and the output attributes, namely, the RGB color and point density. Then the color of each pixel is calculated by querying these attributes along the ray defined by the pixel position and camera's parameters and applying the volume rendering techniques~\cite{max1995optical}. Finally, the radiance field is optimized by supervising rendered color with the ground truth. Once the radiance field is convergent, the images from any view could be synthesized by following the second step mentioned above. 
To better verify the effectiveness, we apply split-layer to three popular NeRF baselines, \textit{i.e.}, the original NeRF~\cite{mildenhall2021nerf} which models the radiance field as a continuous function, the DVGO~\cite{sun2022direct} which adopts a discrete form of direct voxel grid optimization, and DINER~\cite{zhu2023disorder} optimizes with a disorder-invariant hash-table.
Experiments are conduct on the \textit{nerf synthetic} dataset~\cite{mildenhall2021nerf} with a full resolution of $800\times800$, following all authors' default configurations.

\textbf{Results.}
Tab.~\ref{exp-nerf} lists the quantitative comparisons of these six methods. 
Compared with original results, applying split-layer averagely improves the PSNR of NeRF, DVGO, and DINER up to $0.79$dB, $0.22$dB, and $0.27$dB, respectively. 
In addition, Split-enhanced DINER achieves the best results among all the methods. 
These experimental results further verify the generic effectiveness of split-layer.

\section{Conclusion}
We propose \textit{split-layer}, which reformulates the fully-connected layer with high-degree forms.
We theoretically demonstrate that split-layer can expand the feature space to exceedingly high dimensionality, thus significantly enhancing the INR representational capacity without extra parameters.
Extensive experiments validate our theoretical analysis, demonstrating that split-layer can significantly enhance model performance of various INR backbones across various representation and inverse optimization tasks. 
In the future, we will explore other methods to reformulate the neural network and further enlarge the feature space, such as Hilbert kernel or Gaussian kernel.

\clearpage
\bibliography{aaai2026}

\end{document}